\documentclass[smallextended]{svjour3}       
\smartqed  
\usepackage{lineno}
\usepackage[]{hyperref}
\usepackage{array,multirow,graphicx}
\usepackage[numbers]{natbib}
\usepackage[noend]{algpseudocode}
\usepackage{algorithm}
\usepackage{amsmath}
\usepackage{longtable}
\usepackage{subfig}
\usepackage{array,multirow,graphicx}
\begin{document}

\title{Detailed Investigation of Deep Features with Sparse Representation and Dimensionality Reduction in CBIR: A Comparative Study 
}


\author{Ahmad S. Tarawneh         \and
        Ceyhun~Celik      \and 
        Ahmad Hassanat \and
        Dmitry Chetverikov}

\authorrunning{A. S. Tarawneh et al.} 

\institute{Ahmad S. Tarawneh   \and
           Dmitry Chetverikov \at  
		   Department of Algorithms and their Applications\\
			E\"{o}tv\"{o}s Lor\'{a}nd University, Budapest, Hungary\\
			\email{Ahmadtr@caesar.elte.hu , Ahmadtrwh@gmail.com}\\
            \and
            Ceyhun~Celik \at
            Department of Computer Engineering, Gazi University, Ankara, Turkey\\
            \email{celik.ceyhun@gmail.com}\\
            \and
            Ahmad B. Hassanat \at
            Department of Information Technology, Mutah University, Karak, Jordan           
}

\date{Received: date / Accepted: date}

\maketitle

\begin{abstract}
			
	Research on content-based image retrieval (CBIR) has been under development for decades, and numerous methods have been competing to extract the most discriminative features for improved representation of the image content. Recently, deep learning methods have gained attention in computer vision, including CBIR. In this paper, we present a comparative investigation of different features, including low-level and high-level features, for CBIR. We compare the performance of CBIR systems using different deep features with state-of-the-art low-level features such as SIFT, SURF, HOG, LBP, and LTP, using different dictionaries and coefficient learning techniques. Furthermore, we conduct comparisons with a set of primitive and popular features that have been used in this field, including colour histograms and Gabor features. We also investigate the discriminative power of deep features using certain similarity measures under different validation approaches. Furthermore, we investigate the effects of the dimensionality reduction of deep features on the performance of CBIR systems using principal component analysis, discrete wavelet transform, and discrete cosine transform. Unprecedentedly, the experimental results demonstrate high (95\% and 93\%) mean average precisions when using the VGG-16 FC7 deep features of Corel-1000 and Coil-20 datasets with 10-D and 20-D K-SVD, respectively.    
            
\end{abstract}

\section{Introduction}\label{secIntro}
	
	Given a set of images S and an input image i, the goal of a content-based image retrieval (CBIR) system is to search S for i and return the most related/similar images to i, based on their contents. This emergent field responds to an urgent need to search for an image based on its content, rather than typing text to describe image content to be searched for. That is, CBIR systems allow users to conduct a query by image (QBI), and the system's task is to identify the images that are relevant to that image. Prior to CBIR, the traditional means of searching for images was typing a text describing the image content, known as query by text (QBT). However, QBT requires predefined image information, such as metadata, which necessitate human intervention to annotate images in order to describe their contents. This is unfeasible, particularly with the emergence of big data; for example, Flickr creates approximately 3.6 TB of image data, while Google deals with approximately 20,000 TB of data daily\cite{zhang2018survey}, which mostly comprise images and videos. Applications of CBIR are massive in terms of numbers and areas, which include, but are not limited to, medical image analysis \cite{wells2016medical}, image mining\cite{hsu2002image}\cite{sanu2017satellite}\cite{quellec2017deep}, surveillance\cite{hossen2016surveillance},  biometrics\cite{hassanat2014visual}, security\cite{xia2016privacy}\cite{hassanat2017victory}\cite{hassanat2017classification}, and remote sensing\cite{romero2016unsupervised}.
		
	The key to the success of a CBIR system lies in extracting features from an image to define its content. These features are stored to describe each image, which is implemented automatically by the system, using specific algorithms developed for the extraction process. Similarly, a query process is conducted by extracting the same features from the query image to determine the most similar images from a feature dataset, using matching techniques or similarity measures (distance metrics). Therefore, feature extraction is critical for developing an efficient CBIR system. A large number of contributions have been made to obtaining the optimal features that guarantee superior performance, starting from colour histograms \cite{lande2014effective}\cite{hassanat2016fusion}\cite{zhou2018new}, in which the colour frequencies are mainly used to represent the image content. Despite the fact that histograms have been used extensively in CBIR systems, they cannot provide special information regarding the distribution of the colours in the special domain. The co-occurrence matrix has been used to provide such special information in order to gain an improved description of image contents, whereby the appearance of colour intensity with its related neighbours is recorded, followed by the calculation of specific values that are used to describe the contents\cite{srivastava2017content}. Colour co-occurrence matrices are also used to add robustness in describing image contents by extracting different patterns (so-called motifs) \cite{jhanwar2004content}\cite{subrahmanyam2013modified}\cite{elalami2014new} from small blocks in the images. Moreover, in order to colour moments and statistical features, Gabor features \cite{ramasamy2017edge}, wavelet transform\cite{ashraf2018content}, cosine transform \cite{varish2018novel} and Fourier transform \cite{folkers2002content} have been applied to extract different features from images. Furthermore, shape features have been used in CBIR by extracting the main shapes of objects found in the image, and describing them with different shape descriptors, such as Fourier and invariant moments \cite{li2009complex} \cite{wang2011effective}. 
	
	Local feature descriptors (LFDs) or feature points have also been used for CBIR. SIFT \cite{lowe2004distinctive} and SURF\cite{bay2008speeded} are popular methods for extracting feature points to be used in the matching process. The recent and inspiring study \cite{celik2017content} presented a comparison between SIFT and SURF points and investigated the efficiency of these methods compared to a set of other methods, such as the histogram of oriented gradient (HOG), local binary patterns (LBP) and local ternary patterns (LTP). The study proposed a CBIR framework with sparse representation (SR) and covered the performance of these methods using dictionary and coefficient learning, which are the main steps of SR. Three types of dictionary learning methods were used, namely random features, K-means and K-SVD, while the homotopy, lasso, elastic net and iterative shrinkage methods, among others, were used as coefficient learning techniques. The study reported 89\% and 58\% mean average precision (MAP) values for the Coil-20 and Corel-1000 datasets, respectively. 
	
	Recently, efforts have been made to use deep learning (DL) to solve computer vision tasks such as recognition, authentication, segmentation and CBIR \cite{tarawneh2018pilot}\cite{saritha2018content}\cite{tzelepi2018deep}\cite{khatami2018sequential}\cite{pang2018novel}. In general, there are three different means of using deep learning. Firstly, a convolutional neural network (CNN) is trained on a large-scale dataset to use it for classification. Secondly, it is used as transfer learning, where specific layers are weighted from a pre-trained CNN, which is a CNN trained on a large-scale dataset such as ImageNet. Thirdly, the pre-trained CNN is used as a feature extractor, in which case the images will be used as input and the feed-forward will be calculated to extract the features (deep features) from different layers of the CNN models. For CBIR, the CNN can be used as a feature extractor, and the resultant features applied to present the image contents. Although Deep learning is preferred over SR to improve retrieval accuracy in CBIR problems\cite{cheng2016learning}\cite{lei2016learning}\cite{kappeler2016video}\cite{gao2015single}, these algorithms have also been employed together with the same aim \cite{zhao2015heterogeneous}\cite{liu2016robust}\cite{dong2016image}\cite{goh2014learning}. Therefore, this study presents an extensive number of experiments to figure out the best combination between these two leading approaches to miximise the performance of CBIR systems.
	
	Basically, distance metrics and similarity measures play an important role in ensuring the effectiveness of CBIR systems. The significance of this role is evident following extraction of the features from the images, as it is used for finding images whose contents are closer to a query image. In fact, numerous distance metrics have been developed and used for the matching process between a query image and reference images, the most common of which are Euclidian and Manhattan distances, which have been used in various studies\cite{smeulders2000content}. However, in recent years, other measures have been developed mainly to enhance the matching process. For example, a new matching technique to determine the minimum triangular area between a query vector and its relevant images was proposed by\cite{elalami2014new}, and the reported results demonstrated that effective performance can be achieved using this technique. Another dimensionality invariant distance metric known as the Hassanat distance\cite{hassanat2014dimensionality} was proposed to deal with high-dimensional feature vectors, without the need to normalise the data. Practically, many distance metrics are available, which vary in their performance and can be used successfully for different matching tasks, including CBIR\cite{prasath2017distance}.
	
	The goal of this study is to compare the performance of the CBIR system using different features, namely deep features, LFDs and low level features (LLFs). Moreover, we use a SR framework with different dictionaries and coefficient learning methods to investigate the effects of deep features compared to state-of-the-art studies. We also study the enhancement of deep features using discrete cosine transform (DCT)-based coefficients. Finally, we study the effect of dimensionality reduction on the CBIR system performance, using principal component analysis (PCA), discrete wavelet transform (DWT) and DCT with different similarity measures under various validation approaches. 
	
 \textbf{The contributions of this study can be summarized as follows:
	\begin{itemize}
	  \item First of all, different approaches of two leading techniques (DL and SR) are combined to identify the best combination of them, and detailed tests about these combinations are run on two popular data set.
	  \item A large number of experiments (842 different tests) are done to compare the effectiveness of image features (LFDs, Deep Features, ).
	   \item Popular similarity measurements are used to compare the performance of deep features before and after enhancements them using DCT and z-score normalization. 
	   \item Various dimensionality reduction algorithms are employed and tested to investigate the performance of deep features in small features space.
	   \item Our combination of SR with deep features is compared with the state of art methods and shows superior accuracy. 
	\end{itemize}}

	\section{Image features}	\label{secIF}
	CBIR features can be categorised into two types: low-level and high-level features. Low-level features include Gabor features, colour histogram, SIFT, SURF and others, such as those presented in\cite{deselaers2008features} High-level features include deep features extracted from different layers and pre-trained models, such as  AlexNet\cite{krizhevsky2012imagenet}, VGG-16 and VGG-19\cite{simonyan2014very}. In this paper, we compare low-level and high-level features, in addition to comparing the high deep features with one another and investigating the CBIR performance following data pre-processing and dimensionality reduction, as demonstrated in the next sections.	
	
	\subsection{Low-level features} \label{secLLF}
    \subsubsection{Gabor features}\label{secGF}	
	Gabor features are frequently used for different computer vision tasks, including CBIR. In this study, Gabor features are used as in \cite{deselaers2008features}, with different scales and orientations. The 2D Gabor filters in the spatial domain can be defined by
	
    \begin{equation}
		f_{mn}(x,y)  = \frac{1}{2\pi\sigma^2_m}e^{-\frac{x^2+y^2}	     {2\sigma^2_m}}\cos(2\pi(u_{0m}x\cos\theta_n+u_{0m}y\sin\theta_n))
		\label{eq1}
	\end{equation}
		
	where $m$ and $n$ are the scale and orientation of the filters, respectively. The quantity $u_{0m}$ specifies the centre frequency of the filters. The features are extracted by calculating the mean and standard deviation of the images following filtering at five different scales and orientations \cite{squire2000content,park2002fast,zhang2000content}.
	
	\subsubsection{HOG features}\label{secHF}	
	HOG features can be used efficiently for object detection \cite{pang2011efficient,zhu2006fast} and recognition \cite{kyrki2004simple}, in addition to CBIR \cite{desai2017gist}. Typically, the calculation and extraction of these features are carried out as follows.
	The colour and gamma values are normalised as a pre-processing step. Thereafter, the gradient is calculated; generally using horizontal and vertical operators such as $[-1,0,1]$ and $[-1,0,1]^T$. Then, the direction values for each block are calculated and binned in order to eventually extract the HOG features.
		
    \subsubsection{SIFT and SURF}\label{secSS}
	Both SIFT and SURF are reasonably robust LFDs. In SIFT, the features are localised by filtering the image using difference of Gaussians at different scales, following which the local maxima and minima are considered as feature points \cite{lowe2004distinctive}. 
	Speed is a major problem in SIFT; hence, the SURF method was proposed to improve the SIFT method speed by approximating the Laplacian of Gaussian using box filters, which makes the convolution process easily conducted for different scales simultaneously \cite{bay2008speeded}.
	In this paper, HOG, SIFT, SURF, LBP and LTP, among other features, are used for comparison with deep features.
	
	\subsection{High-level features}
	Deep features are those extracted from a specific layer or layers of a pre-trained deep CNN, such as AlexNet. In this work, we extract these features from various layers of different models, namely AlexNet, VGG-16 and VGG-19. Each of these deep models outputs a 4096-dimensional feature vector for each image, which is very high dimensionality, and negatively affects the speed in the matching process. 
	
	\section{Dimensionality reduction}\label{secDimRed}
	In order to alleviate the problem of dimensionality in the deep features, we compare four popular methods that are normally used to reduce the feature space dimensions, namely DCT, PCA, DWT and probability density functions (PDFs).
	
	\subsection{DCT}
	DCT is an invertible linear transform that is widely used in numerous applications and extensively applied to image and audio compression, owing to its ability to extract useful information and exclude redundant data \cite{gupta2012analysis}. A 1D DCT can be defined by
	
    \begin{equation}
		X_k=\sqrt{\frac{2}{N}}\sum_{n=1}^{N} x_n \frac{1}{\sqrt{1+\delta}} \cos\bigg(\frac{\pi}{2N}(n-1)(2k-1)\bigg) , k=1,..., N
		\label{eq2}
	\end{equation}
	
	where $x$ is the input signal, $\delta$ is the Kronecker delta and $N$ is the input signal length. In this work, we reduce the dimensionality of the 1D feature vector extracted from each image by calculating the DCT, and considering the DC coefficient and first $N$ AC coefficients.
	
	\subsection{PCA }
	PCA is a statistical method that makes use of orthogonal transformation to convert a group of variables (in this case, the resultant feature vector) into a group of values known as principal components. Typically, the largest possible data variance is preserved by the first principal component, while the other principal components have different, lower variances. Dimensionality reduction is achieved by maintaining those components with the highest variances, which may explain the main data patterns, and removing those with the lowest variances, which can be considered as redundant data \cite{reich2008principal}.
	
	\subsection{DWT }
\begin{figure}[th]
		\centering
		\includegraphics[width=1\linewidth]{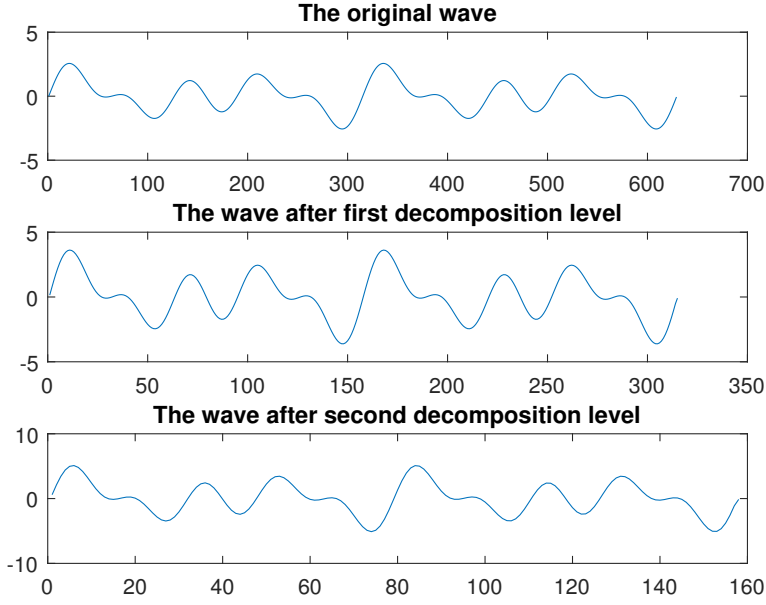}
		\caption[Original hypothetical signal (top), signal after first decomposition level (middle) and the signal after second decomposition level (bottom)]{Original hypothetical signal (top), signal after first decomposition level (middle) and the signal after second decomposition level (bottom)}	
		\label{fig:sin_waves}
	\end{figure}

	DWT has been used extensively in a variety of applications, including dimensionality reduction of the feature vectors of CBIR systems, without a major impact on system performance \cite{rashedi2013simultaneous}. Basically, DWT calculates the approximate coefficients that almost represent the same signal (feature vector) shape. Figure \ref{fig:sin_waves} illustrates a hypothetical signal, in addition to its first and second wavelet decomposition levels. As can be observed from Figure \ref{fig:sin_waves}, we can approximate the signal using 315 or 158 values after the first or second decomposition levels, without excessive loss of its shape and patterns.

    
	We use the Haar DWT to calculate the approximate coefficients, owing to its simplicity and computational efficiency. Algorithm \ref{algo1} defines the steps for calculating the DWT for the dimensionality reduction of our feature vector.
	
	\begin{algorithm}[th]
		\caption{Steps of proposed method for dimensionality reduction}\label{algo1}
		\textbf{Input}: Feature vector X of size N\\
		\textbf{Output}: Coefficient vector X' of size $\approx$ N/2
		\begin{algorithmic}[1]
			\State 	\textbf{for} I=1 to length(X)-1: step 2 \textbf{do}
			\State 	output index $\leftarrow$ $\frac{I+1}{2}$
			\State 	X'(output index)$\leftarrow$$ \frac{ \sum_{c=I}^{I+1} X_{c}        }{\sqrt{2}}$
			\State \textbf{ end for	}
			\State 	\textbf{if} length(X) mod 2 $\neq$ 0 \textbf{then}
			\State 	X' ($\parallel\frac{length(X)}{2}\parallel$)$\leftarrow$ $\frac{ \sum_{1}^{2} X_{length(X)} }{\sqrt{2}}$
			\State 	\textbf{ end if	}
		\end{algorithmic}
	\end{algorithm}
	

	As indicated by the algorithm, each decomposition level reduces the dimensionality of the input feature vector by half.
	
	\subsection{Probability density functions }
	PDF is another technique that can be used to reduce the dimensionality of the feature space \cite{magnatic}. Basically, it depends on calculating the histogram of a group of values within a specific range. The histogram can be converted into a probability density function by
    \begin{equation}
	PDF_i=\frac{F_i}{F_{total}  }
	\label{eq3}
	\end{equation}
	where
	\begin{equation}
	F_{total}=\sum_{i=1}^{N}F_i
    \label{eq4}
	\end{equation}
	
	Here, $F_i$ is the frequency of bin $i$, and $N$ is the number of bins used to build the histogram. It is important to note that the sum of all values of a PDF vector is equal to 1, regardless of the number of bins used.
	\section{Similarity measures}\label{secSimMea}
	As previously mentioned, the similarity measures play a major role in the effectiveness of a CBIR system \cite{hassanat2016fusion}. In this work, we compare the effects of using different similarity measures in CBIR using deep features.
	\subsection{Euclidian distance}
	Euclidian distance (ED) is dominant in this field, owing to its simplicity and common use; however, other metrics tend to perform better, as we will discuss in the experimental section. The ED can be defined by
	
	\begin{equation}
	ED(V1,V2)=\sqrt{\sum_{i=1}^{N}{(V1_i-V2_i)^2} }
    \label{eq5}
	\end{equation}
	where $V1$ and $V2$ are the vectors to be compared and $N$ is the length of each.
	
	\subsection{Manhattan distance}
	
	The Manhattan distance (MD) or city block distance has also been used to compare the feature vectors in CBIR systems. The MD between two vectors is defined by
	
    \begin{equation}
	MD(V1,V2)={\sum_{i=1}^{N}|{(V1_i-V2_i)}| }
    \label{eq6}
	\end{equation}	
	
	\subsection{Hassanat distance}
	The Hassanat distance (HD) is a scale and noise invariant distance metric, where the distance (D) between two points can be defined by
	
	\begin{equation}
	D(V1_i,V2_i)=
	\left\{
	\begin{array}{ll}
	1-\frac{1+min(V1_i,V2_i)}{1+max(V1_i,V2_i)}  & \mbox{, } min(V1_i,V2_i) \geq 0 \\
	1-\frac{1+min(V1_i,V2_i)+|min(V1_i,V2_i)|}{1+max(V1_i,V2_i)+|min(V1_i,V2_i)|}  & \mbox{, } min(V1_i,V2_i) < 0
	\end{array}
	\right.
    \label{eq7}
	\end{equation}
	
	and for the total distance along two vectors 
	\begin{equation}
	HD(V1,V2)={\sum_{i=1}^{N}(V1_i,V2_i) }
    \label{eq8}
	\end{equation}
	
	The advantage of HD is that it is not significantly affected by different data scales, noises and outliers. A careful look at Equation \ref{eq7} reveals that applying this distance to each attribute (dimension) outputs a value within the range of $[0,1]$, where 0 is similar 1 is dissimilar, and in between the similarity is well defined. The value of the distance for each attribute increases logarithmically to reach 1 if the difference reaches infinity. Therefore, if there is an outlier value from noise or a large value from a different scale, regardless of the difference, the maximum addition to the overall distance is 1. In the case of other distances such as MD, if the difference is 100, this number will be added to the overall distance, which allows one feature to dominate the distance. If this is a noise or unscaled datum, we obtain unpredicted results, as the distance becomes biased by large values.
	
	\subsection{Canberra distance}
	Similar to HD, the Canberra distance (CD) does not output more than 1 for each dimension, regardless of the compared values, making it robust to noise and outliers. The CD between two equal-length vectors is defined by
	\begin{equation}
	CD(V1,V2)={\sum_{i=1}^{N}\frac{|V1_i-V2_i|}{|V1_i|+|V2_i|} }
    \label{eq9}
	\end{equation}
	
	However, the CD is not defined when 0 is compared to 0. As the distance between identical values in this metric is 0, we define $CD(0,0) =0$.
	
	\section{Sparse representation }\label{secSR}
	
	Representing signals by means of a simple combination of non-zero elements according to a base is an ancient concept known as the principle of sparsity. SR is based on such a principle, and has been used to solve computer vision problems for the past two decades \cite{zhang2013sparse}. The SR is obtained by solving the following problem
	\begin{equation}
	\min_{\alpha\in R_{}^n} \frac{1}{2} \|x-D\alpha\|_{2}^{2}+\lambda \|\alpha\|_{p}
	\label{SR1}
	\end{equation}
	
	where $x$ is the signal, $D$ is the dictionary, $\alpha$ is the sparse coefficient of signal $x$ and $p$ may be of any value $[0,\infty]$. Dictionary learning and coefficient learning (CL) are the two important steps in SR. While the base vectors are built with the Dictionary learning algorithm, the sparse vector on this base for a given signal is obtained using CL algorithms. K-means and K-singular value decomposition (K-SVD) algorithms are the most widely used Dictionary learning algorithms \cite{lei2014learning}. These algorithms are also known as offline and online techniques, which means that the dictionary is built without sparse coefficients for the former, and the dictionary and coefficients are learned for the latter. As the sparsity term takes numerous parameters, various algorithms have been proposed in the CL step \cite{celik2017content}. However, greedy approaches do not scale effectively for high-dimensional problems, and the results have indicated that iterative-shrinkage algorithms can overcome this problem \cite{zibulevsky2010l1}. The separable surrogate function (SSF) and parallel coordinate descent (PCD) are commonly used algorithms in this class. Furthermore, sequential subspace optimisation (SESOP) speeds up these algorithms, as the process requires a lengthy time in the case of high-dimensional problems \cite{zibulevsky2010l1}.	
	In this study, the K-means and K-SVD algorithms are used to build the dictionary, while the homotopy, lasso, elastic net and SSF are used for the CL step \cite{celik2017content}. For experimental purposes, we divide the Corel-1000 into 100 and 900 images for testing and training, respectively and the Coil-20 into 120 and 1320 images for testing and training, respectively. The ED is used to compare the resultant vectors.
    
 \section{Experimental results and comparisons}\label{secExpR}
	We divide our experiments into two parts. The first part is an investigation into deep features and their performance using different dictionary types of varying sizes, also using CL methods. Moreover, we compare the well-known deep features with state-of-the-art work that has been conducted as a comparative study among SIFT, SURF, HOG, LBP and LTP \cite{celik2017content}. 
	In the second part, we compare different types of deep features obtained from various models to determine how their performance varies with/without pre-processing and dimensionality reduction. Moreover, we compare these deep features with another set of features, including Gabor features, colour histograms, invariant histograms and other techniques, using certain similarity measures with the aforementioned dimensionality reduction methods.
	
\begin{figure}[th]
		\centering
		\includegraphics[width=1\linewidth]{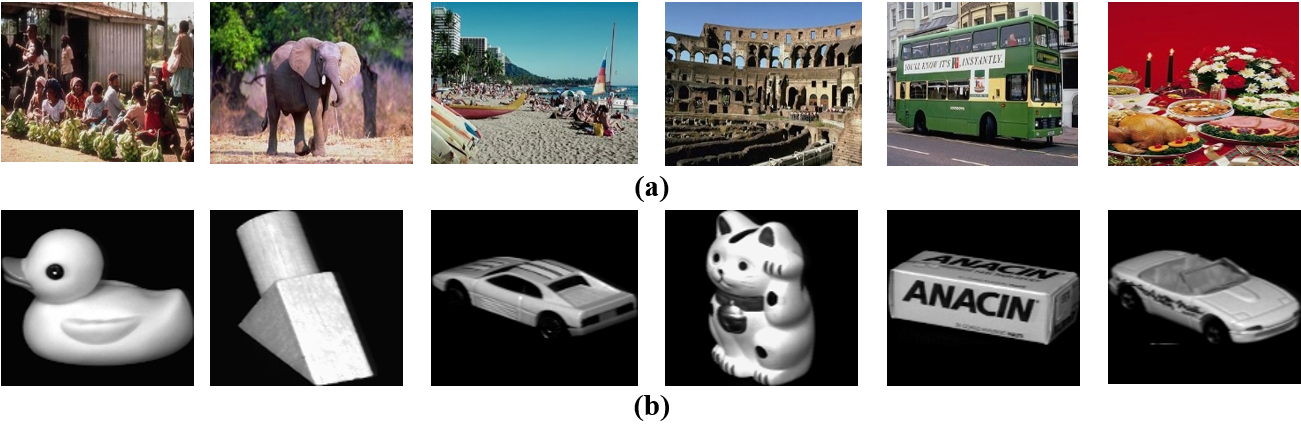}
		\caption[Samples from (a) Corel-1000 and (b) Coil-20 datasets]{Samples from (a) Corel-1000 and (b) Coil-20 datasets}	
		\label{fig:Datasamples}
	\end{figure}

	Similar to the compared studies, we used the Corel-1000 and Coil-20 datasets. Despite the Corel-1000 dataset being relatively old, it is still used in current research because CBIR on this dataset has not yet been perfected. Figure \ref{fig:Datasamples} illustrates samples from both datasets.
	
	We used precision-recall curves and MAP for evaluation of the CBIR system. Precision-recall curve is a commonly used curve to evaluate the data retrieval algorithms. Similarly, the MAP is a single number represents the mean of the precision among a number of query examples and it approximately equals to the area under the precision-recall curve.

	\subsection{Part 1: Sparse representation}
	
    	\begin{table}[th]
		\footnotesize	
		\centering
		\caption{MAP of different features using 512-D K-means on Corel-1000 dataset}			
		\begin{tabular}{l|ccccc|cccccc}			
			\hline
			\hline
			\multicolumn{12}{c}{512-D K-means}\tabularnewline 
			\hline
			& \multicolumn{5}{c|}{Low-level features} & \multicolumn{6}{c}{Deep features}\\
			\hline
			\rotatebox[origin=l]{90}{CL algorithms}&\rotatebox[origin=l]{90}{SIFT}&\rotatebox[origin=l]{90}{SURF}&\rotatebox[origin=l]{90}{HOG}&\rotatebox[origin=l]{90}{LBP}&\rotatebox[origin=l]{90}{LTP}&\rotatebox[origin=l]{90}{AlexNet FC6}&\rotatebox[origin=l]{90}{AlexNet FC7}&\rotatebox[origin=l]{90}{VGG-16 FC6}&\rotatebox[origin=l]{90}{VGG-16 FC7}&\rotatebox[origin=l]{90}{VGG-19 FC6}&\rotatebox[origin=l]{90}{VGG-19 FC7 }\tabularnewline
			\hline			
			Homotopy&0.43&0.40&	0.52&\textbf{0.57}&0.5&0.15&	0.16&0.16&0.18&0.16&0.15\tabularnewline
			Lasso&	0.43&	0.37&	0.50&	0.47&	0.38&	0.16&	0.16&	0.17&	0.16&	0.18&	0.15\tabularnewline
			Elastic net&0.43&0.32&0.49&0.20&0.37&0.14&0.15&0.14&0.14&0.15&	0.15\tabularnewline
			SSF&0.50&0.39&0.44&0.54&0.53&0.44&0.49&0.47&0.50&0.48&	0.51\tabularnewline		
			\hline
			\hline
		\end{tabular}
		\label{table:table1}
	\end{table}	
	
    Tables \ref{table:table1} and \ref{table:table2} display the direct comparison with \cite{celik2017content} when applying the same Dictionary learning and CL methods on the deep features. It can obviously be seen from these tables the results are not satisfying for deep features. HOG and LBP achieved superior results on the Corel-1000 dataset, while LTP recorded the highest MAP rates using all CL methods, except SSF, for both dictionaries on Coil-20. The reason for these results is that, unlike the LFDs, deep features represent an image with one vector, for example. By using a smaller dictionary size, the MAP increased dramatically, as we recorded a 95\% MAP on the Corel-1000 dataset using Homotopy, 10-D K-SVD and VGG-16 FC7 features. Tables \ref{table:table3},\ref{table:table4} and \ref{table:table5} display the MAP values of deep features using different dictionary sizes and CL methods on both datasets.
    
	\begin{table}[th]
		\footnotesize	
		\centering
		\caption{MAP of different features using 512-D K-means and 256-D K-SVD on Coil-20 dataset}			
		\begin{tabular}{c|l|ccccc|cccccc}			
			\hline
			\hline
			& & \multicolumn{5}{c|}{Low-level features} & \multicolumn{6}{c}{Deep features}\\
			\hline
			&\rotatebox[origin=l]{90}{CL algorithms}&\rotatebox[origin=l]{90}{SIFT}&\rotatebox[origin=l]{90}{SURF}&\rotatebox[origin=l]{90}{HOG}&\rotatebox[origin=l]{90}{LBP}&\rotatebox[origin=l]{90}{LTP}&\rotatebox[origin=l]{90}{AlexNet FC6}&\rotatebox[origin=l]{90}{AlexNet FC7}&\rotatebox[origin=l]{90}{VGG-16 FC6}&\rotatebox[origin=l]{90}{VGG-16 FC7}&\rotatebox[origin=l]{90}{VGG-19 FC6}&\rotatebox[origin=l]{90}{VGG-19 FC7 }\tabularnewline
			\hline
			\parbox[t]{2mm}{\multirow{4}{*}{\rotatebox[origin=c]{90}{512-D K-means}}}			
			&Homotopy&0.69&0.53&	0.74&0.74&0.83&	0.34&0.34&0.32&0.33&0.33&	0.34\tabularnewline
			&Lasso&0.48&0.4&	0.63&0.78&0.82&0.35&0.34&0.33&0.33&0.34&	0.34\tabularnewline
			&Elastic net&0.48&0.46&0.63&0.80&0.84&0.12&0.11&0.11&0.11&0.11&	0.12\tabularnewline
			&SSF&0.73&0.57&0.70&0.79&0.80&0.88&0.91&	\textbf{0.92}&\textbf{0.92}&0.9&	0.91\tabularnewline	
			\hline			
			\parbox[t]{2mm}{\multirow{4}{*}{\rotatebox[origin=c]{90}{256-D K-SVD}}}
			&Homotopy&0.75&0.55&0.73&0.75&0.82&0.64&0.66&0.58&0.58&0.61&0.64\tabularnewline
			&Lasso&0.49&0.48&0.65&0.77&0.82&0.62&0.6&0.54&0.54&0.56&0.56\tabularnewline
			&Elastic net&0.49&0.49&0.65&0.78&0.83&0.56&0.56&0.41&0.47&0.52&0.57\tabularnewline
			&SSF&0.71&0.57&0.67&0.79&0.83&0.91&0.91&0.89&0.9&0.91&\textbf{0.92}\tabularnewline	
			\hline
			\hline
		\end{tabular}
		\label{table:table2}
	\end{table}	
    
	\begin{table}[th]
		\footnotesize	
		\centering
		\caption{MAP of deep features using 10-D K-means and 10-D K-SVD on Corel-1000 dataset}			
		\begin{tabular}{l|cccccc|cccccc}			
			\hline
			\hline
			&\multicolumn{6}{c|}{10-D K-means} & \multicolumn{6}{c}{10-D K-SVD}\\
			\hline
			\rotatebox[origin=l]{90}{CL algorithms}&\rotatebox[origin=l]{90}{AlexNet FC6}&\rotatebox[origin=l]{90}{AlexNet FC7}&\rotatebox[origin=l]{90}{VGG-16 FC6}&\rotatebox[origin=l]{90}{VGG-16 FC7}&\rotatebox[origin=l]{90}{VGG-19 FC6}&\rotatebox[origin=l]{90}{VGG-19 FC7 }&\rotatebox[origin=l]{90}{AlexNet FC6}&\rotatebox[origin=l]{90}{AlexNet FC7}&\rotatebox[origin=l]{90}{VGG-16 FC6}&\rotatebox[origin=l]{90}{VGG-16 FC7}&\rotatebox[origin=l]{90}{VGG-19 FC6}&\rotatebox[origin=l]{90}{VGG-19 FC7 }\tabularnewline
			\hline			
			Homotopy&0.81&0.8&0.84&0.81&0.73&0.79&0.62&0.84&0.87&\textbf{0.95}&0.82&0.90\tabularnewline
			Lasso&0.80&\textbf{0.86}&0.85&0.76&0.74&0.80&0.59&0.83&0.85&0.94&0.77&0.89\tabularnewline
			Elastic net&0.76&0.85&0.80&0.78&0.73&0.81&0.60&0.82&0.82&0.93&0.76&0.88\tabularnewline
			SSF&0.81&0.80&0.85&0.83&0.75&0.80&0.63&0.83&0.88&0.94&0.82&0.88\tabularnewline			
			\hline
			\hline
		\end{tabular}
		\label{table:table3}
	\end{table}	
	
	\begin{table}[th]
		\footnotesize	
		\centering
		\caption{MAP of deep features with different dictionary sizes using K-means on Corel-1000 and Coil-20 datasets}			
		\begin{tabular}{c|l|cccccc|cccccc}			
			\hline
			\hline
			& &\multicolumn{6}{c|}{Corel-1000} & \multicolumn{6}{c}{Coil-20}\\
			\hline
			&\rotatebox[origin=l]{90}{CL algorithms}&\rotatebox[origin=l]{90}{AlexNet FC6}&\rotatebox[origin=l]{90}{AlexNet FC7}&\rotatebox[origin=l]{90}{VGG-16 FC6}&\rotatebox[origin=l]{90}{VGG-16 FC7}&\rotatebox[origin=l]{90}{VGG-19 FC6}&\rotatebox[origin=l]{90}{VGG-19 FC7 }&\rotatebox[origin=l]{90}{AlexNet FC6}&\rotatebox[origin=l]{90}{AlexNet FC7}&\rotatebox[origin=l]{90}{VGG-16 FC6}&\rotatebox[origin=l]{90}{VGG-16 FC7}&\rotatebox[origin=l]{90}{VGG-19 FC6}&\rotatebox[origin=l]{90}{VGG-19 FC7 }\tabularnewline
			\hline			
			\parbox[t]{2mm}{\multirow{4}{*}{\rotatebox[origin=c]{90}{20-D}}}
			&Homotopy&0.69&0.64&0.78&0.62&0.79&0.67&0.87&0.9&0.87&0.88&0.88&0.91\tabularnewline
			&Lasso&0.79&0.80&0.86&0.86&0.88&0.85&0.83&0.85&0.84&0.8&0.82&0.88\tabularnewline
			&Elastic net&0.80&0.82&0.85&\textbf{0.89}&0.87&0.86&0.82&0.86&0.87&0.83&0.83&0.89\tabularnewline
			&SSF&0.75&0.74&0.84&0.77&0.83&0.77&0.89&0.91&0.90&0.91&0.9&\textbf{0.93}\tabularnewline			
			\hline
			\parbox[t]{2mm}{\multirow{4}{*}{\rotatebox[origin=c]{90}{30-D}}}			
			&Homotopy&0.60&	0.48&0.70&0.51&0.63&0.55&0.85&0.89&0.85&0.87&0.88&0.89\tabularnewline
			&Lasso&0.70&0.72&0.82&0.79&0.80&0.80&0.86&0.88&0.83&0.85&0.89&0.9\tabularnewline
			&Elastic net&0.75&0.75&0.83&0.82&\textbf{0.84}&0.82&0.86&0.87&0.85&0.88&0.91&0.92\tabularnewline
			&SSF&0.68&0.72&0.82&0.73&0.79&0.76&0.9&0.93&0.9&0.91&0.92&\textbf{0.93}\tabularnewline
			\hline			
			\parbox[t]{2mm}{\multirow{4}{*}{\rotatebox[origin=c]{90}{40-D}}}
			&Homotopy&0.55&0.46&0.67&0.47&0.62&0.51&0.87&0.84&0.87&0.85&0.86&0.87\tabularnewline
			&Lasso&0.67&0.68&0.77&0.72&0.73&0.75&0.86&0.84&0.87&0.87&0.86&0.89\tabularnewline
			&Elastic net&0.68&0.70&0.77&0.74&0.74&\textbf{0.78}&0.87&0.86&0.9&0.88&0.89&0.91\tabularnewline
			&SSF&0.69&0.66&0.78&0.72&0.73&0.74&0.91&0.9&\textbf{0.92}&\textbf{0.92}&\textbf{0.92}&\textbf{0.92}\tabularnewline
			\hline
			\parbox[t]{2mm}{\multirow{4}{*}{\rotatebox[origin=c]{90}{50-D}}}
			&Homotopy&0.51&0.47&0.63&0.42&0.58&0.42&0.82&0.85&0.82&0.83&0.85&0.8\tabularnewline
			&Lasso&0.59&0.66&0.72&0.66&0.70&0.65&0.84&0.87&0.83&0.87&0.85&0.82\tabularnewline
			&Elastic net&0.59&0.67&\textbf{0.76}&0.67&0.72&0.71&0.88&0.89&0.89&0.88&0.88&0.86\tabularnewline
			&SSF&0.63&0.69&0.74&0.64&0.71&0.68&0.90&0.91&0.90&\textbf{0.93}&\textbf{0.93}&0.91\tabularnewline			
			\hline
			\hline
		\end{tabular}
		\label{table:table4}
	\end{table}	
	
	\begin{table}[th]
		\footnotesize	
		\centering
		\caption{MAP of deep features with different dictionary sizes using K-SVD on Corel-1000 and Coil-20 datasets}			
		\begin{tabular}{c|l|cccccc|cccccc}			
			\hline
			\hline
			& &\multicolumn{6}{c|}{Corel-1000} & \multicolumn{6}{c}{Coil-20}\\
			\hline
			&\rotatebox[origin=l]{90}{CL algorithms}&\rotatebox[origin=l]{90}{AlexNet FC6}&\rotatebox[origin=l]{90}{AlexNet FC7}&\rotatebox[origin=l]{90}{VGG-16 FC6}&\rotatebox[origin=l]{90}{VGG-16 FC7}&\rotatebox[origin=l]{90}{VGG-19 FC6}&\rotatebox[origin=l]{90}{VGG-19 FC7 }&\rotatebox[origin=l]{90}{AlexNet FC6}&\rotatebox[origin=l]{90}{AlexNet FC7}&\rotatebox[origin=l]{90}{VGG-16 FC6}&\rotatebox[origin=l]{90}{VGG-16 FC7}&\rotatebox[origin=l]{90}{VGG-19 FC6}&\rotatebox[origin=l]{90}{VGG-19 FC7 }\tabularnewline
			\hline			
			\parbox[t]{2mm}{\multirow{4}{*}{\rotatebox[origin=c]{90}{20-D}}}
			&Homotopy&0.81&0.82&0.84&0.88&0.90&0.87&0.84&0.87&0.83&0.89&0.88&0.87\tabularnewline
			&Lasso&0.78&0.79&0.82&0.86&0.88&0.87&0.80&0.81&0.78&0.82&0.84&0.82\tabularnewline
			&Elastic net&0.75&0.82&0.83&\textbf{0.91}&0.86&0.89&0.79&0.84&0.81&0.86&0.86&0.82\tabularnewline
			&SSF&0.79&0.73&0.82&0.82&0.86&0.80&0.88&0.91&0.87&\textbf{0.93}&0.91&0.91\tabularnewline			
			\hline
			\parbox[t]{2mm}{\multirow{4}{*}{\rotatebox[origin=c]{90}{30-D}}}			
			&Homotopy&0.75&0.80&0.86&0.81&0.76&0.82&0.86&0.87&0.9&0.84&0.9&0.91\tabularnewline
			&Lasso&0.70&0.77&0.87&0.80&0.74&0.82&0.82&0.84&0.89&0.82&0.88&0.87\tabularnewline
			&Elastic net&0.75&0.79&\textbf{0.89}&0.86&0.77&0.84&0.82&0.86&0.87&0.83&0.85&0.89\tabularnewline
			&SSF&0.73&0.72&0.83&0.75&0.75&0.75&0.9&0.9&0.91&0.91&0.92&\textbf{0.93}\tabularnewline
			\hline			
			\parbox[t]{2mm}{\multirow{4}{*}{\rotatebox[origin=c]{90}{40-D}}}
			&Homotopy&0.75&0.80&0.84&0.80&0.80&0.78&0.86&0.84&0.86&0.88&0.89&0.91\tabularnewline
			&Lasso&0.71&0.77&0.85&0.80&0.79&0.76&0.82&0.82&0.83&0.85&0.88&0.89\tabularnewline
			&Elastic net&0.77&0.77&\textbf{0.87}&0.84&0.80&0.78&0.78&0.82&0.84&0.87&0.88&0.9\tabularnewline
			&SSF&0.70&0.74&0.78&0.75&0.77&0.71&0.89&0.9&0.9&0.92&0.92&\textbf{0.93}\tabularnewline
			\hline
			\parbox[t]{2mm}{\multirow{4}{*}{\rotatebox[origin=c]{90}{50-D}}}
			&Homotopy&0.76&0.72&0.81&0.74&0.76&0.74&0.87&0.81&0.87&0.87&0.86&0.86\tabularnewline
			&Lasso&0.73&0.69&0.81&0.71&0.76&0.74&0.85&0.78&0.86&0.85&0.85&0.86\tabularnewline
			&Elastic net&0.78&0.73&\textbf{0.85}&0.75&0.79&0.75&0.86&0.81&0.84&0.88&0.87&0.87\tabularnewline
			&SSF&0.75&0.68&0.76&0.69&0.74&0.67&0.90&0.90&\textbf{0.92}&\textbf{0.92}&0.91&0.91\tabularnewline		
			\hline
			\hline
		\end{tabular}
		\label{table:table5}
	\end{table}
	
	A dictionary of size 10 could not be built on the Coil-20 dataset because the number of classes was 20. The superior results achieved on the Coil-20 dataset reached 93\% and were achieved by VGG-19 FC7 and SSF using both dictionaries. In general, the features extracted using the VGG-16 model appear to have achieved superior performance. The use of elastic net and SSF achieved the best MAP, using both dictionaries with different sizes. Tables \ref{table:table6} and \ref{table:table7} display the MAP averages for all CL and features used, on both datasets and for all dictionary sizes. This aids in determining which features provide superior performance, and which is the most suitable CL method.
	
	\begin{table}[th]
		\footnotesize
		\centering
		\caption{Average MAP values of all CL used for all dictionary sizes}		
		\begin{tabular}{cccc}
			\hline
			\hline		
			Homotopy&Lasso&Elastic net&SSF\tabularnewline
			\hline
			0.786&0.807&0.819&0.825\tabularnewline
			\hline
			\hline
		\end{tabular}
		\label{table:table6}
	\end{table}
	
	\begin{table}[th]
		\footnotesize
		\centering
		\caption{Average MAP values of all features used for all dictionary sizes}		
		\begin{tabular}{cccccc}
			\hline
			\hline		
			AlexNet FC6&AlexNet FC7&VGG-16 FC6&VGG-16 FC7&VGG-19 FC6&VGG-19 FC7\tabularnewline
			\hline
			0.755&0.801&0.837&0.827&0.805&0.825\tabularnewline
			\hline
			\hline
		\end{tabular}
		\label{table:table7}
	\end{table}
	
	As both tables indicate, on average, the features extracted from VGG-16 achieved superior performance, while using SSF as the CL provided the highest MAP rate.

	\subsection{Part 2: Similarity measures}
	
	In this part, we used the Corel-1000 and Coil-20 datasets with leave-one-out cross-validation to obtain valid comparisons. Leave-one-out cross-validation is an effective means of determining the CBIR system performance, as it uses all images in the dataset as query images, which mimics a real-world application. However, this approach is not used on large datasets, because it requires a very long time, particularly for a large number of features.
	Table \ref{table:table8} (a) displays the MAP values of all deep features using the aforementioned distance metrics without performing DTC pre-processing or dimensionality reduction. As can be observed from these results, the CD achieved superior performance, followed by HD; this is because neither metrics were affected by noise and outliers, as explained previously. The best MAP reached 84.2\%, recorded by using features extracted from VGG-16 FC6 on the Corel dataset and 90.1\% using VGG-19 FC6 and CD. The MD and ED exhibited almost the same performance for all features, while HD was slightly superior to both.
	
	DCT is normally used for dimensionality reduction, on either one or two dimensions. It has recently been revealed that deep features provide better results after its representation in DCT domain \cite{ghosh2016deep}. In this study, the MAP of the CBIR using different similarity measures was significantly enhanced after applying 1D DCT, owing to the strong energy-compaction characteristic of the DCT. However, we used DCT without removing any coefficients, simply for processing the signal in a manner that enhances the features for improved recognition. Unexpectedly, this is proven to be suitable, perhaps owing to the cosine function used in DCT, as it scales or transforms the data so as to enhance matching.

	\begin{table}[th]
		\footnotesize	
		\centering
		\caption{MAP of different deep features using various similarity measures}			
		\begin{tabular}{c|l|cccc|cccc}			
			\hline
			\hline
			& &\multicolumn{4}{c|}{Coil-20} & \multicolumn{4}{c}{Corel-1000}\\
			\hline
			&NET&HD&CD&MD&ED&HD&CD&MD&ED\tabularnewline
			\hline
			\multicolumn{10}{c}{(a)} \\
			\hline			
			\parbox[t]{2mm}{\multirow{6}{*}{\rotatebox[origin=c]{90}{Without DCT}}}
			&AlexNet FC6&0.863&0.864&0.860&0.858&0.73&0.752&0.717&0.717\tabularnewline
			&AlexNet FC7&0.861&0.862&0.861&0.863&0.691&0.716&0.682&0.688\tabularnewline
			&VGG-16 FC6&0.885&0.889&0.880&0.881&0.774&\textbf{0.842}&0.758&0.758\tabularnewline
			&VGG-16 FC7&0.875&0.887&0.876&0.879&0.768&0.815&0.755&0.758\tabularnewline
			&VGG-19 FC6&0.895&\textbf{0.901}&0.892&0.892&0.777&0.841&0.756&0.756\tabularnewline
			&VGG-19 FC7&0.883&0.895&0.884&0.887&0.761&0.810&0.746&0.749\tabularnewline	
			\hline
			\multicolumn{10}{c}{(b)} \\
			\hline
			\parbox[t]{2mm}{\multirow{6}{*}{\rotatebox[origin=c]{90}{With DCT}}}			
			&AlexNet FC6&0.862&0.859&0.859&0.858&0.769&0.778&0.696&0.717\tabularnewline
			&AlexNet FC7&0.874&0.874&0.870&0.863&0.756&0.792&0.706&0.688\tabularnewline
			&VGG-16 FC6&0.890&0.893&0.884&0.881&0.821&0.853&0.728&0.758\tabularnewline
			&VGG-16 FC7&0.897&0.902&0.892&0.879&0.816&0.862&0.771&0.758\tabularnewline
			&VGG-19 FC6&0.900&0.902&0.894&0.892&0.823&0.854&0.726&0.756\tabularnewline
			&VGG-19 FC7&0.902&\textbf{0.906}&0.898&0.887&0.809&\textbf{0.863}&0.760&0.749\tabularnewline
			\hline
			\multicolumn{10}{c}{(c)} \\
			\hline			
			\parbox[t]{2mm}{\multirow{6}{*}{\rotatebox[origin=c]{90}{Z-score normalization}}}
			&AlexNet FC6&0.863&0.869&0.859&0.858&0.722&0.786&0.685&0.678\tabularnewline
			&AlexNet FC7&0.874&0.882&0.869&0.868&0.734&0.801&0.699&0.692\tabularnewline
			&VGG-16 FC6&0.888&0.897&0.884&0.884&0.762&0.862&0.716&0.707\tabularnewline
			&VGG-16 FC7&0.897&0.904&0.893&0.892&0.799&0.872&0.764&0.757\tabularnewline
			&VGG-19 FC6&0.898&0.903&0.894&0.893&0.762&0.864&0.714&0.705\tabularnewline
			&VGG-19 FC7&0.903&\textbf{0.909}&0.898&0.897&0.792&\textbf{0.873}&0.753&0.746\tabularnewline
			\hline		
			\multicolumn{10}{c}{(d)} \\
			\hline	
			\parbox[t]{2mm}{\multirow{6}{*}{\rotatebox[origin=c]{90}{10 PCA}}}
			&AlexNet FC6&0.721&0.797&0.838&0.844&0.443&0.641&0.776&0.792\tabularnewline
			&AlexNet FC7&0.715&0.811&0.841&0.846&0.479&0.654&0.741&0.752\tabularnewline
			&VGG-16 FC6&0.791&0.848&0.865&0.859&0.589&0.740&0.830&0.846\tabularnewline
			&VGG-16 FC7&0.766&0.830&0.852&0.855&0.608&0.741&0.833&0.847\tabularnewline
			&VGG-19 FC6&0.793&0.844&0.869&\textbf{0.873}&0.568&0.752&0.850&\textbf{0.863}\tabularnewline
			&VGG-19 FC7&0.720&0.794&0.851&0.865&0.622&0.744&0.829&0.842\tabularnewline
			\hline		
			\multicolumn{10}{c}{(e)} \\
			\hline	
			\parbox[t]{8mm}{\multirow{6}{*}{\rotatebox[origin=c]{90}{\begin{tabular}{c@{}c@{}} 10 PCA \& \\ Z-score normalization \end{tabular}}}}
			&AlexNet FC6&0.809&0.798&0.819&0.814&0.665&0.641&0.734&0.739\tabularnewline
			&AlexNet FC7&0.821&0.811&0.836&0.835&0.670&0.654&0.737&0.751\tabularnewline
			&VGG-16 FC6&0.866&0.848&0.869&0.868&0.759&0.741&0.814&0.822\tabularnewline
			&VGG-16 FC7&0.841&0.830&0.858&0.862&0.771&0.741&0.832&\textbf{0.850}\tabularnewline
			&VGG-19 FC6&0.852&0.844&0.867&\textbf{0.870}&0.783&0.752&0.840&\textbf{0.850}\tabularnewline
			&VGG-19 FC7&0.825&0.795&0.854&0.867&0.770&0.745&0.826&0.838\tabularnewline
			\hline			
			\multicolumn{10}{c}{(f)} \\
			\hline
			\parbox[t]{2mm}{\multirow{6}{*}{\rotatebox[origin=c]{90}{10 bins PDFs}}}
			&AlexNet FC6&0.178&0.171&0.178&0.181&0.119&0.117&0.119&0.119\tabularnewline
			&AlexNet FC7&0.211&\textbf{0.213}&0.210&0.212&0.127&0.121&0.127&0.127\tabularnewline
			&VGG-16 FC6&0.166&0.151&0.165&0.167&0.138&0.130&0.139&0.140\tabularnewline
			&VGG-16 FC7&0.154&0.144&0.153&0.156&0.135&0.132&0.135&0.135\tabularnewline
			&VGG-19 FC6&0.179&0.179&0.179&0.181&0.145&0.134&0.145&\textbf{0.146}\tabularnewline
			&VGG-19 FC7&0.193&0.193&0.192&0.193&0.137&0.133&0.137&0.138\tabularnewline
			\hline
			\hline
		\end{tabular}
		\label{table:table8}
        \end{table}
        \clearpage
          
	\begin{figure}[th]
    \centering
		\includegraphics[width=1\linewidth]{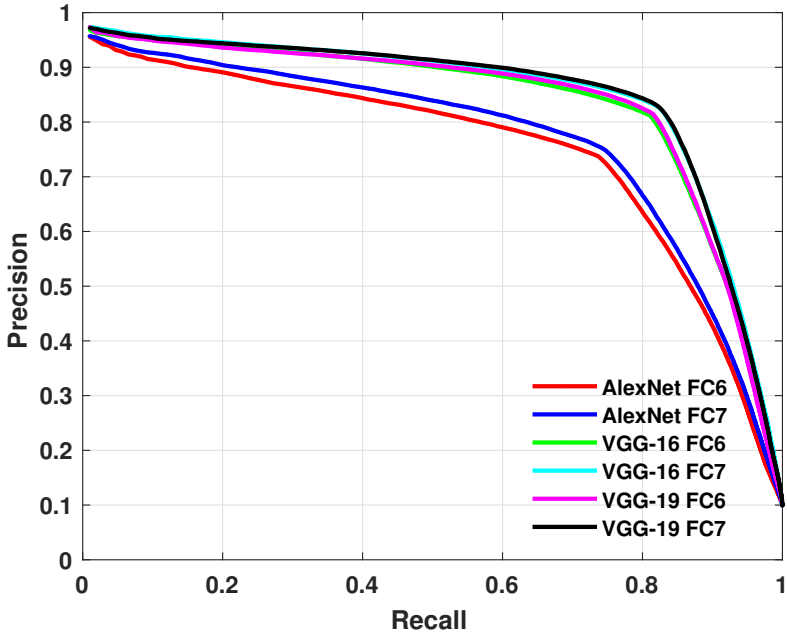}
		\caption[Precision-recall curves of deep features with DCT processing and normalisation using CD on Corel-1000 database]{Precision-recall curves of deep features with DCT processing and normalisation using CD on Corel-1000 database}	
		\label{fig:corel}		
	\end{figure}
    
	A closer inspection of the results in Table \ref{table:table8} (b) reveals that the potential exists for deep features to be enhanced by pre-processing techniques such as the DCT for improved retrieval MAP. Such an enhancement was significant when using HD and CD compared to the original deep features, which were extracted directly from the deep models (CNNs). The best MAP values following DCT processing reached 86.3\% and 90.6\% on the Corel-1000 and Coil-20 datasets, respectively, using the CD and VGG-19 FC7 features.  However, In the case of MD, there was no significant improvement, while the MAP remained the same when using ED following pre-processing. It is also interesting to note that HD benefited from the DCT more than CD did, owing to the nature of CD, as most of the distances between different values were equal to 1, while this was not the case in HD.
    
    The signal following DCT pre-processing contains different scaled features, and as we conducted similarity matching based on distance metrics, there might be a risk of false decisions by allowing large-scale features to dominate the final distance. Therefore, we opted for data normalisation. Table \ref{table:table8} (c) displays the MA§P values following Z-score normalisation of the DCT coefficients.
	
\begin{figure}[th]
    \centering
		\includegraphics[width=1\linewidth]{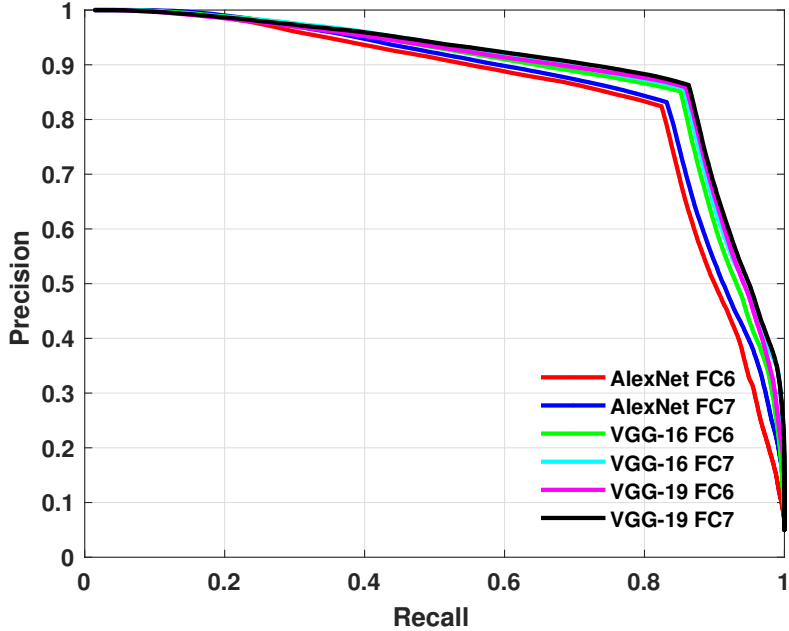}
		\caption[Precision-recall curves of deep features with DCT processing and normalisation using CD on Coil-20 database]{Precision-recall curves of deep features with DCT processing and normalisation using CD on Coil-20 database}	
		\label{fig:coil}		
	\end{figure}

	As can be seen from Table \ref{table:table8} (c), the Z-score normalisation of the DCT coefficients enhanced certain results, as the best MAP increased to 87.3\% and 90.9\% on Corel-1000 and Coil-20, respectively.
	Figures \ref{fig:corel} and \ref{fig:coil} depict the precision-recall curves of different pre-processing and normalisation of the deep features using CD on both datasets.

	Normally, the number of deep features is relatively high, and when using similarity measures, the matching process becomes very slow; therefore, reducing the number of features will enhance the searching speed for query images. It is important in this case to reduce the number of features without affecting the CBIR system accuracy. Reduction techniques such as DCT, PCA or DWT significantly reduce the dimensionality of the feature vectors, while maintaining effective system performance. 
    
\begin{figure}[th]
    \centering
		\includegraphics[width=1\linewidth]{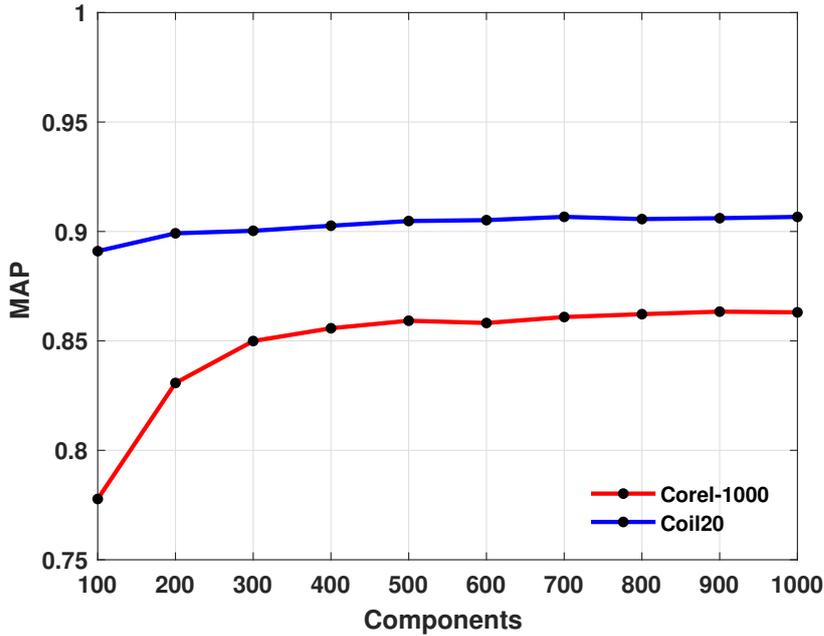}
		\caption[MAP values using different number of DCT coefficients on both datasets]{MAP values using different number of DCT coefficients on both datasets}	
		\label{fig:redudct}		
	\end{figure}

	Based on Table \ref{table:table8} (c), we selected the first N coefficients to approximate the original signal to a smaller number of coefficients. Figure \ref{fig:redudct} illustrates the effect of the number of DCT coefficients used on the CBIR system MAP.
    
	Intuitively, using more DCT coefficients results in a higher MAP being obtained; however, a closer look at Figure \ref{fig:redudct} reveals that the MAP becomes extremely close to that of the original signal when using only 300 coefficients on both datasets, which allows for faster CBIR without affecting the retrieving results.  
	PCA reduces the dimensionality more effectively than DCT, by obtaining almost the same results with a significantly lower number of features. For example, Table \ref{table:table8} (d) illustrates the MAP using only 10 principal components (those with the highest data variance) with different distance metrics, while Table \ref{table:table8} (e) illustrates the CBIR MAP using the same number of components after normalising the data.
	
	As can be seen from Tables \ref{table:table8} (d) and \ref{table:table8} (e), the MAP is still high after transforming the features from 4096 to 10-dimensions. In this case, ED and MD perform better, owing to the small number of features, which reduces the probability of features that dominate the distance.
	
	We opted to use the 10 components of PCA for the experiments displayed in Table \ref{table:table8} (e), because using these achieves the highest system MAP, as indicated in Figure 6.

    \begin{figure}[th]
    \centering
		\includegraphics[width=1\linewidth]{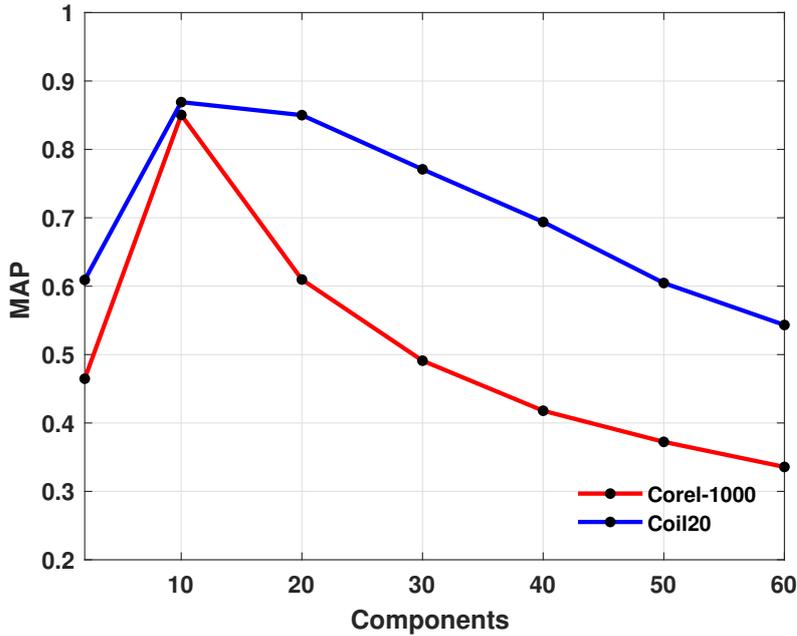}
		\caption[MAP as a function of different PCA components]{MAP as a function of different PCA components}	
		\label{fig:redupca}		
	\end{figure}
	
	In order to investigate the effect of the dimensionality reduction of the deep features on the CBIR system MAP further, we used the DWT for dimensionality reduction, as in \cite{agarwal2014content,reduction}. We applied three decomposition levels, and at each DWT level, the number of features was reduced by half. Table \ref{table:table9} illustrates the system performance after three decomposition levels using the VGG-16 FC6 and VGG-19 FC6 with CD (with the best features and distance metric indicated in Table \ref{table:table8} (a)) on both datasets.
	
	\begin{table}[th]
		\footnotesize
		\centering
		\caption{MAP and number of features for each DWT level}		
		\begin{tabular}{cccc}
			\hline
			\hline		
			Level&Number of features&MAP of Corel-1000&MAP of Coil-20\tabularnewline
			\hline
			Without DWT&4096&0.842&0.901\tabularnewline
			1&2048&0.80&0.894\tabularnewline
			2&1024&0.734&	0.879\tabularnewline
			3&512&0.662&0.848\tabularnewline
			\hline
			\hline
		\end{tabular}
		\label{table:table9}
	\end{table}
	
	As can be observed from Table \ref{table:table9}, the signal or feature vector began to change as the number of levels used increased, as the MAP deceased by approximately 7\% after each level on the Corel-1000 dataset, which is a significant system performance degradation. In the case of the Coil-20 dataset, the loss was not significant; however, the CBIR system performance with dimensionality reduction using DWT was not satisfactory if compared to DCT and PCA, in terms of the number of features and system MAP.  
	Converting the features into PDF reduces the dimensionality by grouping the features in the same range in order to produce less length feature vectors \cite{magnatic}. Table \ref{table:table8} (f) presents the MAP of the 10 bins PDF using the studied distances.
	
	It can be noted from Table \ref{table:table8} (f) that the results are not satisfactory. Perhaps grouping the same range deep features together in a bin reduces the significance of the features by being in their original place. Moreover, binning does not transform the features in a specific manner within a space, but simply performs blind grouping without prior information regarding the features, while it may be effective for special features as in \cite{magnatic}, it appears not to with complex structure features such as deep features. 
	The term error rate (ER) is used as an evaluation measure in this comparison, and is given by $1-p(1)$, where $p(1)$ is the precision at the first retrieved image. In general, a lower error rate means that the relevant images are retrieved earlier. Table \ref{table:table10} displays the best MAP and ER values for different deep features (all deep features using CD with DCT and normalisation, as in Table \ref{table:table8} (c)) compared to the low-level features presented in \cite{deselaers2008features}.

	\begin{table}[th]
		\footnotesize
		\centering
		\caption{MAP and ER of different deep features compared to the low-level features presented in \cite{deselaers2008features}on the Corel-1000 dataset}		
		\begin{tabular}{lcc}
			\hline
			\hline		
			Features&MAP&ER\tabularnewline
			\hline
			Colour histogram&50.5&16.9\tabularnewline
			LF SIFT global search&38.3&37.2\tabularnewline
			LF patches histogram&48.3&17.9\tabularnewline
			LF SIFT histogram&48.2&25.6\tabularnewline
			Inv. feature histogram (monomial)&47.6&19.2\tabularnewline
			MPEG7: scalable color&46.7&25.1\tabularnewline
			LF patches signature&40.4&24.3\tabularnewline
			Gabor histogram&41.3&30.5\tabularnewline
			32 x 32 image&37.6&47.2\tabularnewline
			MPEG7: color layout&41.8&35.4\tabularnewline
			X x 32 image&24.3&55.9\tabularnewline
			Tamura texture histogram&38.2&28.4\tabularnewline
			LF SIFT signature&36.7&35.1\tabularnewline
			Grey value histogram&31.7&45.3\tabularnewline
			LF patches global&30.5&42.9\tabularnewline
			MPEG7: edge histogram&40.8&32.8\tabularnewline
			Inv. feature histogram (relational)&34.9&38.3\tabularnewline
			Gabor vector&23.7&65.5\tabularnewline
			Global texture feature&	26.3&51.4\tabularnewline
			AlexNet FC6&78.6&4.4\tabularnewline
			AlexNet FC7&80.1&4.3\tabularnewline
			VGG-16 FC6&86.2&3.2\tabularnewline
			VGG-16 FC7&87.2&\textbf{2.6}\tabularnewline
			VGG-19 FC6&86.4&\textbf{2.6}\tabularnewline
			VGG-19 FC7&\textbf{87.3}&2.8\tabularnewline
			\hline
			\hline
		\end{tabular}
		\label{table:table10}
	\end{table}
	\clearpage
    
\begin{figure}[th]
    \centering
		\includegraphics[width=1\linewidth]{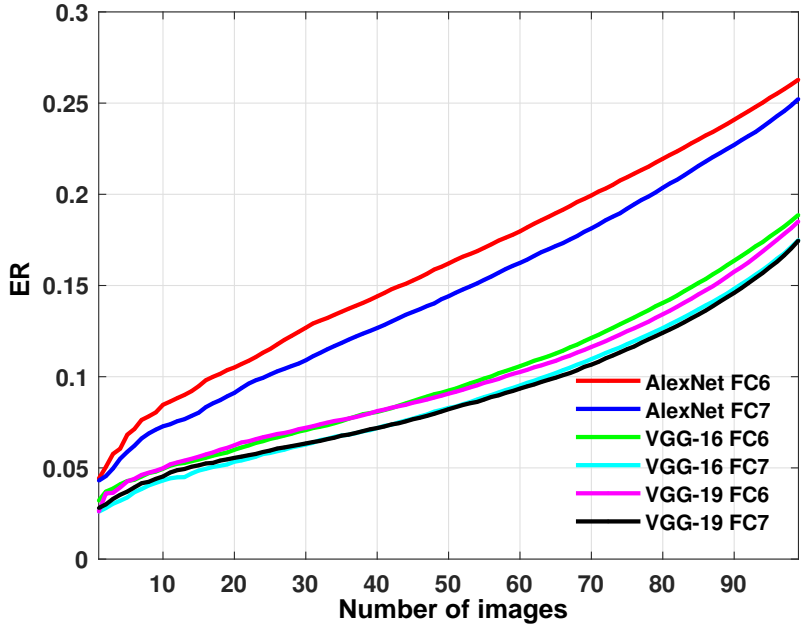}
		\caption[ER of all deep features following each retrieved image]{ER of all deep features following each retrieved image}	
		\label{fig:ER}		
	\end{figure}

	Figure \ref{fig:ER} presents the ER following each retrieved image up to 99, which is the number of relevant images of each image in the Corel-1000 dataset. 
        
	As can be seen from Figure \ref{fig:ER}, very deep features such as VGG-19 FC7 yield a lower ER than less deep features such AlexNet FC6. In VGG models, the relevant images are retrieved earlier than in the AlexNet feature.

    
	\section{Conclusion}
	In this paper, different types of low-level and deep features for CBIR have been tested and compared. The comparison was conducted using different SR and CL methods, and various similarity measures with different validation approaches. Furthermore, we examined the features with/without data normalization and experimentally demonstrated the effect of different dimensionality reduction techniques on the system performance, using two popular data sets in this field. 
	
Eight hundred forty-two different tests are done on both Corel and Coil datasets using different SR methods, dictionary learning algorithms, similarity measures, dimensional reduction techniques and many deep features from different layers deep models.
	
Results show that combination of DL and SR lead to accurate retrieval models. Especially, when the results of these combinations are compared with LFD based SR, the usage of DL as deep features increase the retrieval accuracy.

The experimental results indicate high MAPs on both datasets using K-SVD and homotopy, particularly when using a small dictionary size. However, in general, SSF achieved the best results, while the VGG-16 features were optimum in the SR framework. Despite the fact that SSF is rarely used in CBIR systems, in this study, it is proven that the SSF overcomes the other common CL algorithms. Therefore testing SSF with different surrogate functions is recommended for future studies.

We determined that the system performance varies based on the used distance and pre-processing methods. In general, the VGG models performed better in extracting more efficient features. 

Selecting the optimal distance relies on the data itself, and in this study, CD and HD achieved superior performance to MD and ED in most cases. However, in the case of PCA reduction, ED provided superior results to the other metrics. Furthermore, the results indicate that the deep features can be enhanced by using DCT as a pre-processing stage. Such a step increases the system performance and achieves higher accuracy than using the pure deep features; however, selecting the appropriate distance metric is critical to the system performance following pre-processing.

Moreover, PCA was found to be the most suitable option among the investigated dimensionality reduction algorithms, as it reduced the dimensionality of the deep features dramatically, to 10 features, while maintaining effective performance. Similarly, the DCT provides a high approximation of the performance with the original features, by using only 300 features.

In future studies, We will investigate the performance of SSF with deep features in many other computer vision problems, including face recognition, fingerprints identification and authentication, facial and medical image retrieval, etc.

\begin{acknowledgements}
The first author would like to thank Tempus Public Foundation for sponsoring his PhD study, also, this paper is under the project EFOP-3.6.3-VEKOP-16-2017-00001 (Talent Management in Autonomous Vehicle Control Technologies), and supported by the Hungarian Government and co-financed by the European Social Fund.
\end{acknowledgements}


\bibliographystyle{unsrtnat}
\bibliography{References}

\end{document}